\pdfoutput=1
\documentclass[11pt]{article}

\usepackage[final]{acl}

\usepackage{times}
\usepackage{latexsym}

\usepackage[T1]{fontenc}

\usepackage[utf8]{inputenc}

\usepackage{microtype}

\usepackage{inconsolata}

\usepackage{graphicx}
\graphicspath{{figs/}}

\usepackage{amsmath}
\usepackage{amssymb}
\usepackage{bm}
\usepackage{threeparttable}
\usepackage{booktabs}
\usepackage{multicol}
\usepackage{multirow}
\usepackage{array}

\usepackage{xcolor,pifont}
\usepackage{bbding}
\usepackage{colortbl}

\usepackage{arydshln}

\usepackage{makecell}

\usepackage{subfig}
\usepackage[ruled,linesnumbered,boxed]{algorithm2e}
\usepackage{hyperref}

\definecolor{deepred}{rgb}{0.698,0.133,0.133}
\definecolor{blue}{rgb}{0,0,1}

\definecolor{myred}{rgb}{1.0, .349, .3686}
\definecolor{CoINBlue}{rgb}{0.129, 0.6196, 0.7373}     
\definecolor{CoINRed}{rgb}{0.8392, 0.1569, 0.1569}   
\definecolor{shadecolor}{rgb}{0.92,0.92,0.92}

\title{Enhancing Multimodal Continual Instruction Tuning with BranchLoRA}

\author{Duzhen Zhang\textsuperscript{1}\footnotemark[1]\footnotemark[2], Yong Ren\textsuperscript{2}\footnotemark[1], Zhong-Zhi Li\textsuperscript{2}, Yahan Yu\textsuperscript{3}, Jiahua Dong\textsuperscript{1}, Chenxing Li\textsuperscript{4}\\ \textbf{Zhilong Ji\textsuperscript{5}} \and \textbf{Jinfeng Bai\textsuperscript{5}}\\
  \textsuperscript{1}Mohamed bin Zayed University of Artificial Intelligence, Abu Dhabi, UAE\\
\textsuperscript{2}Institute of Automation, Chinese Academy of Sciences, Beijing, China\\
\textsuperscript{3}Kyoto University, Kyoto, Japan \textsuperscript{4}Tencent AI Lab, Beijing, China\\
\textsuperscript{5}Tomorrow Advancing Life, Beijing, China\\
\texttt{duzhen.zhang@mbzuai.ac.ae}, \texttt{\{thurenyong, dongjiahua1995\}}\texttt{@gmail.com}\\
\texttt{lizhongzhi2022@ia.ac.cn}, \texttt{yahan@nlp.ist.i.kyoto-u.ac.jp}, \texttt{chenxingli@tencent.com}\\
             }

\begin{document}
\maketitle
\renewcommand{\thefootnote}{\fnsymbol{footnote}}
\footnotetext[1]{Equal contributions.}
\footnotetext[2]{Corresponding author.}
\renewcommand{\thefootnote}{\arabic{footnote}}

\begin{abstract}

Multimodal Continual Instruction Tuning (MCIT) aims to finetune Multimodal Large Language Models (MLLMs) to continually align with human intent across sequential tasks. 
Existing approaches often rely on the Mixture-of-Experts (MoE) LoRA framework to preserve previous instruction alignments. 
However, these methods are prone to Catastrophic Forgetting (CF), as they aggregate all LoRA blocks via simple summation, which compromises performance over time.
In this paper, we identify a critical parameter inefficiency in the MoELoRA framework within the MCIT context. 
Based on this insight, we propose BranchLoRA, an asymmetric framework to enhance both efficiency and performance. 
To mitigate CF, we introduce a flexible tuning-freezing mechanism within BranchLoRA, enabling branches to specialize in intra-task knowledge while fostering inter-task collaboration. 
Moreover, we incrementally incorporate task-specific routers to ensure an optimal branch distribution over time, rather than favoring the most recent task. 
To streamline inference, we introduce a task selector that automatically routes test inputs to the appropriate router without requiring task identity.
Extensive experiments on the latest MCIT benchmark demonstrate that BranchLoRA significantly outperforms MoELoRA and maintains its superiority across various MLLM sizes.\footnote{Our code will be available at \url{https://github.com/BladeDancer957/BranchLoRA}.}

\end{abstract}

\section{Introduction}

Multimodal Large Language Models (MLLMs) \cite{DBLP:conf/icml/0008LSH23,liu2023llava,bai2023qwen}, which combine a visual encoder with a LLM, have achieved remarkable success in addressing various multimodal tasks.  
Instruction tuning \cite{DBLP:journals/corr/abs-2305-06500} plays a pivotal role in aligning MLLMs with human intent, enabling the creation of versatile models with general-purpose capabilities.  
In practical scenarios, MLLMs are often required to adapt to new instructions to support evolving functionalities as knowledge and societal needs advance \cite{zheng2024beyond}.  
However, current MLLMs remain static, limiting their ability to accommodate continually emerging demands. Retraining MLLMs from scratch to meet these requirements is costly and inefficient. 
To address this challenge, recent research has framed the problem within the paradigm of Multimodal Continual Instruction Tuning (MCIT). MCIT seeks to continually finetune MLLMs for new tasks while preserving their strong performance on previously learned ones.

\begin{figure}[t!]
\centering
  \includegraphics[width=1\linewidth]{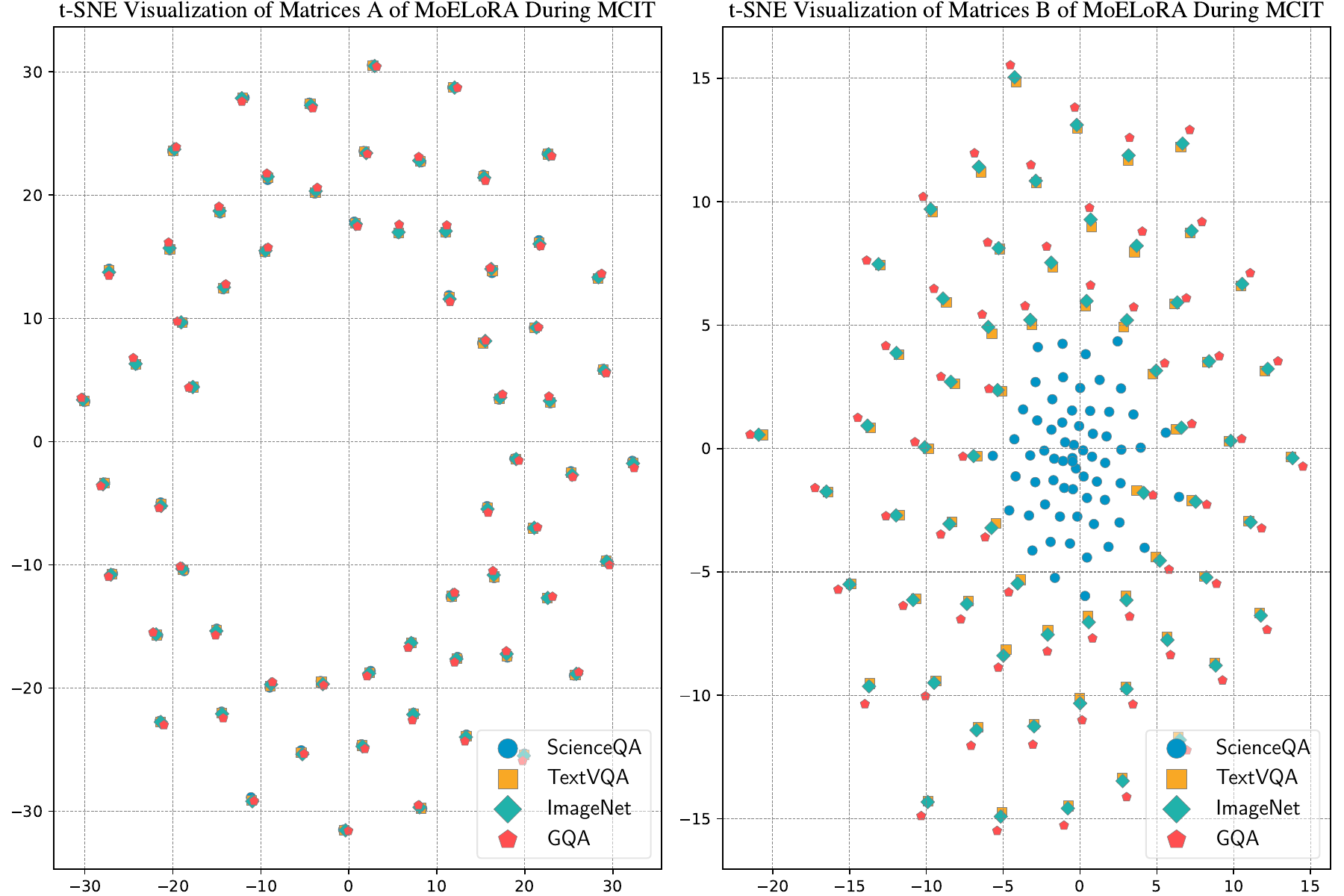}
    \caption{MoELoRA parameter analysis during MCIT across $4$ sequential tasks: matrices = layers $\times$ experts.}
\label{intro:exp}
\end{figure}

\begin{figure}[t!]
\centering
  \includegraphics[width=1\linewidth]{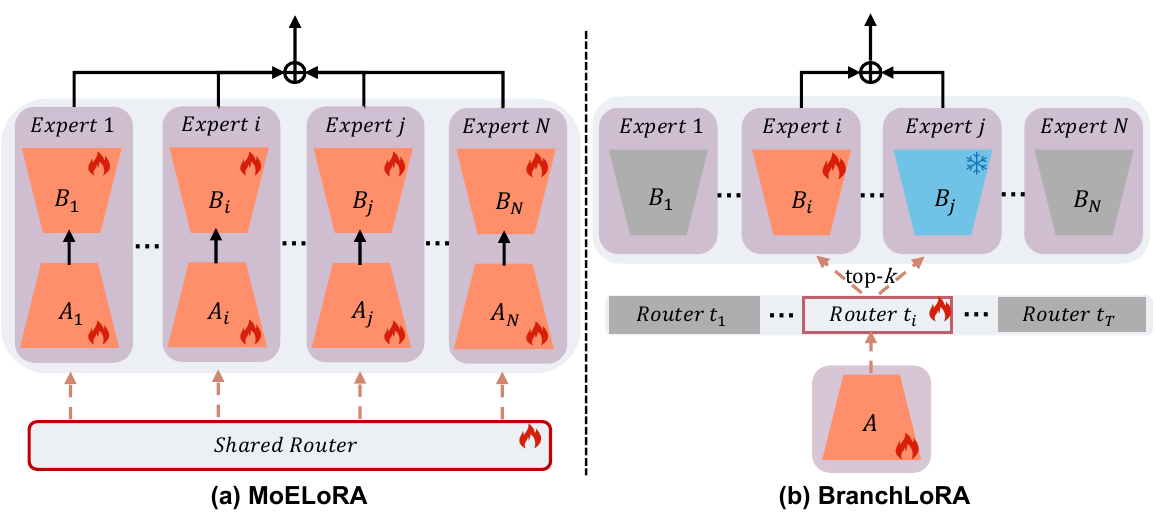} 
    \caption{Diagram of MoELoRA and BranchLoRA.}
\label{intro:lora}
\end{figure}

Multimodal Continual Instruction Tuning (MCIT) faces a significant challenge: Catastrophic Forgetting (CF), where models lose or overwrite previously acquired knowledge when adapting to new tasks \cite{mccloskey1989catastrophic,zhang2023continual,10323204}.  
To mitigate this issue, Mixture-of-Experts (MoE) LoRA \cite{hulora,chen2024coin}, illustrated in Figure \ref{intro:lora} (a), utilizes multiple specialized experts (\emph{i.e.}, LoRA blocks) to capture distinct knowledge from sequential tasks while employing a shared router to modulate their contributions.  
However, experiments on the MCIT benchmark have revealed a critical limitation of MoELoRA related to parameter inefficiency. As visualized in Figure \ref{intro:exp}, we analyzed the parameter behavior of MoELoRA when finetuned continually on $4$ sequential tasks.  
The results highlight a key observation: the parameters in matrices $\bm{A}$ of MoELoRA tend to converge, capturing shared patterns across all tasks, whereas the parameters in matrices $\bm{B}$ remain distinct, focusing on the unique aspects. This suggests that MoELoRA suffers from parameter redundancy. Based on these findings, we propose that an improved architecture should adopt an asymmetric structure to better balance task-shared and task-specific learning.

Moreover, as shown in Figure \ref{intro:lora} (a), MoELoRA aggregates all experts during MCIT, making it susceptible to overwriting previously learned knowledge when adapting to new tasks. Additionally, the router in MoELoRA is shared across tasks, and its continuous updates optimize the expert distribution primarily for the most recent task. This results in the forgetting of distributions optimized for earlier tasks.  
In summary, these two limitations cause MoELoRA to remain vulnerable to CF.

To address the above limitations, we introduce BranchLoRA (shown in Figure \ref{intro:lora} (b)), an asymmetric framework designed to enhance MCIT performance. In this design, the shared matrix $\bm{A}$ serves as a ``tree trunk'' capturing task-invariant patterns, while multiple matrices $\bm{B}$ (\emph{i.e.}, experts) act as ``branches'' that encode task-specific knowledge for sequential tasks. Instead of aggregating all experts, BranchLoRA adopts a dynamic sparse selection strategy, selecting only the top-$k$ experts based on their distribution. 
To further enhance intra-task learning, foster inter-task collaboration, and reduce inter-task interference during MCIT, we introduce a flexible tuning-freezing mechanism. 
When learning a new task, the router accesses frozen experts to leverage transferable knowledge from previous tasks, while optimizing tunable experts to acquire task-specific information for the new task.

To prevent bias toward the most recent task and maintain an optimal expert distribution over time, BranchLoRA incrementally incorporates task-specific routers. Furthermore, we integrate a task selector that automatically routes test samples to the appropriate router without requiring explicit task identity during inference, ensuring greater alignment with real-world scenarios. 
These innovations collectively strengthen BranchLoRA's anti-forgetting capabilities and deliver a marked improvement in overall MCIT performance.

Our contributions can be summarized as follows:

\begin{itemize}

\item Through MCIT experiments, we identify parameter inefficiency in MoELoRA and propose an asymmetric BranchLoRA architecture to balance shared and task-specific learning.

\item We introduce a flexible tuning-freezing mechanism and task-specific routers within BranchLoRA, allowing experts to capture intra-task knowledge, foster inter-task collaboration, and minimize inter-task interference, thereby mitigating CF more effectively.

\item Extensive experiments on the recent MCIT benchmark demonstrate that BranchLoRA significantly outperforms MoELoRA. Furthermore, BranchLoRA consistently achieves superior performance across different MLLM sizes (\emph{e.g.}, LLaVA-7B and LLaVA-13B).

\end{itemize}

\section{Related Work}

\subsection{MLLMs}

Recently, MLLMs have seen significant advancements, extending LLMs to handle visual and textual inputs \cite{zhang2024mm,chen2023vlp,zhao2025chartcoder,zhao2025chartedit}. They utilize the inherent reasoning abilities of LLMs alongside the high-quality representations of visual foundation models to achieve complex multimodal reasoning \cite{li2025system} and content comprehension. BLIP2 \cite{DBLP:conf/icml/0008LSH23} bridges the gap between image and text by integrating a frozen LLM with a visual tower, using the Q-Former projector to enable modality alignment. LLaVA \cite{liu2023llava} and MiniGPT4 \cite{zhu2023minigpt} simplify the alignment process with a straightforward linear projector, applying instruction tuning to better align with human intent. Recent models, such as LLaVA-1.5 \cite{liu2023improved}, ShareGPT4V \cite{chen2023sharegpt4v}, and LLaVA-NeXT \cite{liu2024llavanext}, have further refined these strategies, enhancing performance across various multimodal benchmarks. These innovations underscore the growing capabilities of MLLMs in tackling complex reasoning tasks.

\subsection{MCIT}

While MLLMs have made significant progress, regular updates are crucial to endow them with new capabilities and ensure they stay aligned with the rapidly evolving landscape of human knowledge \cite{wu2024continual,zheng2024towards,zheng2025lifelong}. To achieve this, MCIT is essential, as it allows models to continuously incorporate emerging data without the costly need for retraining from scratch. Recently, MCIT benchmarks have been introduced to finetune MLLMs across sequential tasks \cite{he2023continual, chen2024coin}. However, MCIT faces the challenge of CF, where the model forgets earlier knowledge when learning new tasks \cite{goodfellow2013empirical,dong2022federated_FCIL,dong2023federated,zhang2023task,zhai2023investigating,zhai2024investigating,zhang2024balancing,yu2024flexible,zhang2025federated}. Existing approaches, such as the MoELoRA paradigm, attempt to maintain prior instruction alignment \cite{chen2024coin}, but they still struggle with forgetting due to the simple aggregation of all LoRA experts, leading to performance degradation over time.

\section{Preliminary}

\subsection{Problem Formulation}

MCIT \cite{he2023continual,chen2024coin} is designed to enable MLLMs to continually instruction-tune on new datasets without incurring the expense of full re-training. Unlike traditional continual learning \cite{de2021continual}, MCIT emphasizes the effective use of natural language instructions to mitigate CF and promote knowledge transfer. It is formulated as a sequential stream of datasets, denoted by $\mathcal{T}_\text{seq} = \{t_1, ..., t_T\}$, where $T$ represents the total number of datasets or tasks. Importantly, these datasets are diverse in nature and are not confined to specific domains or categories. 

Each dataset $t_i \in \mathcal{T}_\text{seq}$ includes a natural language instruction $I_{t_i}$, a training set $\mathcal{D}_{t_i}^\text{train}$, and a test set $\mathcal{D}_{t_i}^\text{test}$. The objective of MCIT is to sequentially train a single model $\mathcal{M}$ on the dataset stream while maintaining robust performance across all previously encountered tasks. Notably, during inference, the model is presented with test samples without prior knowledge of their associated tasks.

\subsection{MoELoRA}

To alleviate the challenge of CF, \citeauthor{chen2024coin} adopt the widely used MoELoRA approach \cite{dou2023loramoe,liu2023moelora} within the context of MCIT. This method leverages multiple experts to acquire specialized knowledge tailored to different tasks. MoELoRA is composed of two key components: a fixed expert pool and a router. The pool consists of multiple identical yet independent LoRA blocks \cite{hulora}, which are prepended into each Feed-Forward (FF) layer of MLLM. Meanwhile, the router is responsible for modeling a probability distribution that determines the output weights of these experts \cite{fedus2022switch}. In particular, given an intermediate representation $\bm{x}$ produced by the preceding attention layer, the output of the MoELoRA layer can be expressed as follows:
\begin{equation}
\begin{aligned}
    &\bm{h} = \bm{x}\bm{W}_f + \frac{\alpha}{r}\sum_{j=1}^N R(\bm{x})_{[j]}E_j(\bm{x})\\
    &R(\bm{x}) = \textit{Softmax}(\bm{x}_{[0]}\bm{W}_r)\text{,}\ E_j(\bm{x}) = \bm{x}\bm{A}_j\bm{B}_j
\end{aligned}\text{,}
\label{eq:moelora}
\end{equation}
where $\bm{W}_f\in\mathbb{R}^{d_{\text{in}} \times d_{\text{out}}}$ represents the parameters of the FF layer, while the rank $r$ determines the number of trainable parameters. The constant hyperparameter $\alpha$ is introduced to finetune the effect of $r$. The model employs $N$ experts, where the router $R(\cdot)$ takes the first token of an intermediate representation $\bm{x}_{[0]}\in\mathbb{R}^{1\times d_{\text{in}}}$ as input and is parameterized by trainable $\bm{W}_r\in\mathbb{R}^{d_{\text{in}} \times N}$ to assign output weights across these experts. Each expert, denoted as $E_j(\cdot)$, is characterized by two trainable low-rank matrices, $\bm{A}_j\in\mathbb{R}^{d_{\text{in}} \times \frac{r}{N}}$ and $\bm{B}_j\in\mathbb{R}^{\frac{r}{N} \times d_{\text{out}}}$. These matrices have a reduced rank of $\frac{r}{N}$, ensuring that the total number of trainable parameters matches that of a single LoRA setup, thereby maintaining computational efficiency.

\begin{figure*}[t!]
\centering
  \includegraphics[width=1\linewidth]{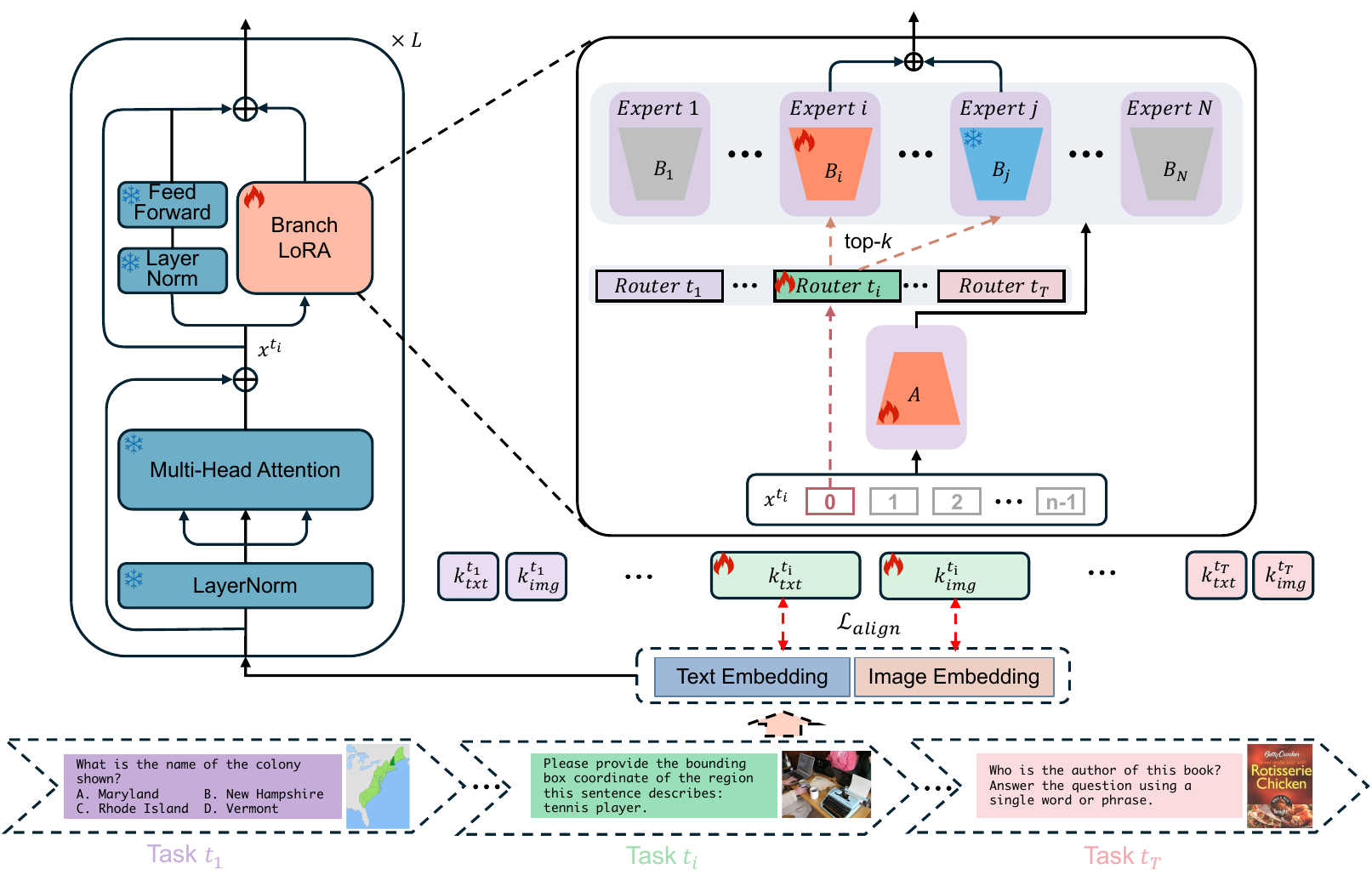} 
\caption{The overall framework of the proposed BranchLoRA. The shared matrix $\bm{A}$ captures task-invariant patterns, while multiple matrices $\bm{B}$ (\emph{i.e.}, experts) encode task-specific knowledge for sequential tasks. BranchLoRA is integrated into the feed-forward module of each MLLM layer, with its input being the intermediate representation $\bm{x}^{t_i}$ from the multi-head attention module. The task-specific router processes the first token of $\bm{x}^{t_i}$ to generate expert weights, enabling a dynamic sparse selection strategy to combine expert outputs. Additionally, task-specific keys are trained to facilitate automatic task selection, eliminating the need for explicit task identity during inference.}
\label{fig:main}
\end{figure*}

\section{Method}

\subsection{Analysis of MoELoRA in MCIT Context}

The motivation behind MoELoRA \cite{chen2024coin} is to leverage multiple experts (\emph{i.e.}, smaller LoRA blocks) to capture specialized knowledge from different tasks. To explore how these multiple experts reduce task interference, we conduct experiments by continually finetuning MoELoRA on $4$ sequential tasks from the MCIT benchmark \cite{chen2024coin}: ScienceQA \cite{lu2022learn}, TextVQA \cite{singh2019towards}, ImageNet \cite{deng2009imagenet}, and GQA \cite{hudson2019gqa}. As illustrated in Figure \ref{intro:exp}, we use the t-SNE technique \cite{van2008visualizing} to visualize the parameters of matrices $\bm{A}$ and $\bm{B}$ across all experts from every MLLM layer. This analysis reveals a key finding:

\textbf{Observation:} \textit{When multiple experts are finetuned continually on sequential tasks, the parameters of matrix $\bm{A}$ tend to converge, while those of matrix $\bm{B}$ remain distinguishable.}

Specifically, all matrices $\bm{A}$ exhibit strong similarity, leading to noticeable overlaps across four sequential tasks, as illustrated in Figure~\ref{intro:exp}. In contrast, matrices $\bm{B}$ are more distinct and easier to differentiate. We hypothesize that this discrepancy arises from their initialization methods: matrix $\bm{A}$ tends to capture shared features across tasks, whereas matrix $\bm{B}$ focuses on adapting to task-specific variations. This suggests that the existing MoELoRA approach may suffer from parameter redundancy in the context of MCIT. A similar pattern was also reported by HydraLoRA \cite{tian2024hydralora} in multi-task finetuning scenarios.

\subsection{BranchLoRA}

Building on the previous observation, we propose an asymmetric BranchLoRA framework aimed at improving both efficiency and performance. In this framework, \textbf{the parameters of matrix $\bm{A}$ are shared} across sequential tasks to optimize parameter usage, while multiple matrices $\bm{B}$ are employed to capture task-specific knowledge. To further enhance intra-task learning, promote inter-task collaboration, and reduce inter-task interference during MCIT, we introduce a flexible tuning-freezing mechanism and task-specific routers within BranchLoRA. This approach effectively mitigates CF. The overview of the BranchLoRA framework is shown in Figure \ref{fig:main}.

\subsubsection{Flexible Tuning-freezing Mechanism}

Rather than always aggregating all experts, which can lead to overwriting previously learned knowledge when adapting to new tasks, we employ a sparse selection strategy. This approach dynamically selects the top-$k$ experts based on their probability distribution. The router $R(\cdot)$ in equation (\ref{eq:moelora}) is modified as follows:
\begin{equation}
R(\bm{x}) = \textit{Softmax}(\text{top-}k(\bm{x}_{[0]}\bm{W}_r))\text{,}
\label{eq:router}
\end{equation}
where $\text{top-}k(\cdot)$ selects the $k$ most relevant experts and assigns a value of $-\infty$ to the others.

\begin{figure}[t!]
\centering
  \includegraphics[width=1\linewidth]{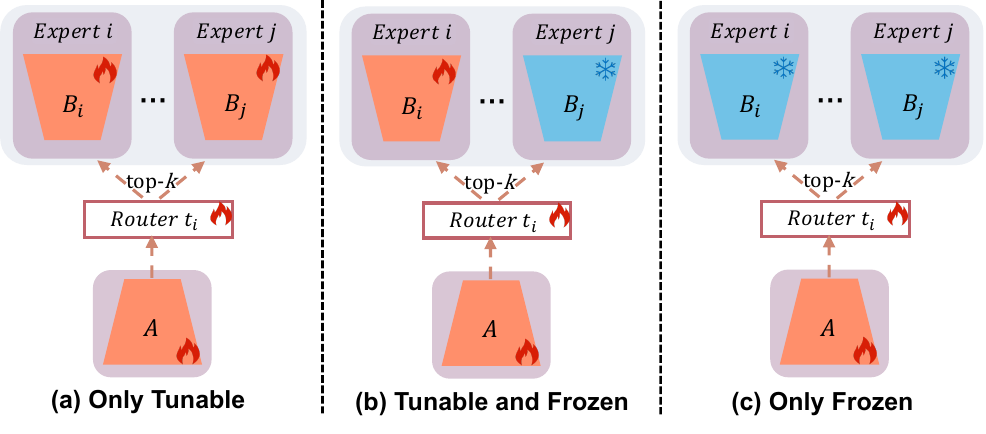} 
\caption{Diagram of the Flexible Tuning-Freezing Mechanism. $3$ distinct expert combinations for an input during new task $t_i$ training: (a) Only Tunable, (b) Tunable and Frozen, and (c) Ony Frozen, with the shared matrix $\bm{A}$ and task-specific Router $t_i$ being tunable.}
\label{fig:tuning}
\end{figure}

To further enhance experts (\emph{i.e.}, matrices $\bm{B}$) with both intra-task knowledge and inter-task collaboration, we introduce a flexible tuning-freezing mechanism. To be more specific, after training on the current task, we analyze the distribution of router outputs across all samples. The top-$k$ most activated experts are then frozen during training on subsequent tasks to preserve the task-specific knowledge of the current task. In this way, when confronted with a new task, the router can access these frozen experts to leverage transferable knowledge from previous tasks, while optimizing the unfrozen experts to capture task-specific information for the new task. As shown in Figure \ref{fig:tuning}, during new task training, the router can activate (a) only the tunable experts, (b) both tunable and previously frozen experts, or (c) exclusively the frozen experts from past tasks (with only the matrix A and router being tunable). This mechanism enables experts to collaboratively consolidate their knowledge, mirroring how the human brain strengthens and integrates new information with existing memories.

\subsubsection{Task-specific Routers with Auto-Selector}

As tasks are trained sequentially, the router in Equation (\ref{eq:router}) is updated continuously, which can result in the forgetting of distributions optimized for earlier tasks. To avoid bias toward the most recent task and ensure an optimal expert distribution over time, we incrementally introduce task-specific routers. The updated version of Equation (\ref{eq:router}) is as follows:
\begin{equation}
R^{t_i}(\bm{x}) = \textit{Softmax}(\text{top-}k(\bm{x}^{t_i}_{[0]}\bm{W}_r^{t_i}))\text{,}
\end{equation}
where the task-specific router $R^{t_i}(\cdot)$ with trainable $\bm{W}_r^{t_i}$ is utilized to assign output weights to different experts by using the first token of an intermediate representation $\bm{x}^{t_i}_{[0]}$ from task $t_i$ as input.

However, task-specific routers rely on task identity during inference, which poses challenges in real-world scenarios where task identity may not always be available. To overcome this limitation, inspired by \cite{wang2022learning, wang2022dualprompt}, we propose learning task-specific keys to enable automatic task selection without requiring explicit task identity during inference. These keys are progressively aligned with the image and text embeddings of samples from task $t_i$ during training. The approach is formulated as follows:
\begin{equation}
\begin{aligned}
    \mathcal{L}_{\text{align}} &= \sum\nolimits_{j} (1-Cos(\bm{e}^{t_i}_{j,\text{img}}, \bm{k}^{t_i}_{\text{img}})) \\ 
    &+ \sum\nolimits_j (1-Cos(\bm{e}^{t_i}_{j,\text{txt}}, \bm{k}^{t_i}_{\text{txt}}))
\end{aligned}\text{,}
\end{equation}
where $Cos(\cdot)$ represents cosine similarity, while $\bm{e}^{t_i}_{j,\text{img}}$ and $\bm{e}^{t_i}_{j,\text{txt}}$ denote the image and text embeddings of the $j$-th sample from task $t_i$, respectively. $\bm{k}^{t_i}_{\text{img}}$ and $\bm{k}^{t_i}_{\text{txt}}$ represent the trainable keys corresponding to the images and texts in task $t_i$.

During inference, we can compute the similarity between the embeddings of a test sample and the trained keys for each task, and automatically route the sample to the router corresponding to the keys with the highest similarity.

Finally, the total loss in BranchLoRA is:
\begin{equation}
    \mathcal{L}_{\text{total}} = \mathcal{L}_{t_i} + \lambda \mathcal{L}_{\text{align}}\text{,}
\end{equation}
where $\mathcal{L}_{t_i}$ is the autoregressive generation loss for task $t_i$ and $\lambda$ is the loss coefficient.

\section{Experimental Settings}

\subsection{Datasets}

We adopt the setup of the latest MCIT benchmark, CoIN \cite{chen2024coin}, which incorporates $8$ multimodal datasets spanning various domains and tasks: ScienceQA \cite{lu2022learn}, TextVQA \cite{singh2019towards}, ImageNet \cite{deng2009imagenet}, GQA \cite{hudson2019gqa}, VizWiz \cite{gurari2018vizwiz}, Grounding \cite{kazemzadeh2014referitgame,mao2016generation}, VQAv2 \cite{goyal2017making}, and OCR-VQA \cite{mishra2019ocr}. 
These datasets are used in the same training order as in the main experiment. 
Detailed statistics for these datasets are provided in Table \ref{aptable:datasets} of Appendix \ref{ap:datasets}.

\begin{table*}[t]
	\renewcommand\arraystretch{1.3}
	\renewcommand\tabcolsep{4.0pt}
	\centering
	\resizebox{\linewidth}{!}{
		\begin{tabular}{ccccccccc|ccc}
			\toprule[1.2pt]
			 \multirow{2}{*}{\textbf{Method}} & \multicolumn{8}{c|}{\textbf{Accuracy on Each Task}} & \multicolumn{3}{c}{ \textbf{Overall Results}} \\ \cline{2-12}
             
			& ScienceQA & TextVQA & ImageNet & GQA & VizWiz & Grounding & VQAv2 & OCR-VQA & ACC$\Uparrow$ & MAA$\Uparrow$ &BWT$\Uparrow$ \\ \hline
			
            Zero-shot & 49.91 & 2.88 & 0.33 & 2.08 & 0.90 & 0.00 & 0.68 & 0.17 & -- &7.12 & --  \\ \cdashline{1-12} 
			
			\multirow{2}{*}{\makecell[c]{LoRA}} & 82.45 & 49.99 & 96.05 & 56.40 & 55.45 & 31.27 & 62.20 & 57.08 & \multirow{2}{*}{28.74} &\multirow{2}{*}{32.97} & \multirow{2}{*}{-32.62} \\
            
			& 21.26 & 28.74 & 10.25 & 36.78 & 32.45 & 0.83 & 42.50 & 57.08  \\ \cdashline{1-12}

            	\multirow{2}{*}{\makecell[c]{$\text{LwF}^{*}$}} & 81.36 & 50.59 & 96.84 & 51.98 & 48.19 & 25.13 & 41.30 & 64.12 & \multirow{2}{*}{30.41}&\multirow{2}{*}{34.95} & \multirow{2}{*}{-27.03} \\ 
			& 26.78 & 37.52 & 12.64 & 35.18 & 25.24 & 2.87 & 38.92 & 64.12  \\ \cdashline{1-12}

            	\multirow{2}{*}{\makecell[c]{$\text{EWC}^{*}$}} & 82.81 & 51.76 & 96.80 & 46.19 & 48.68 & 26.82 & 66.37 & 63.46 & \multirow{2}{*}{32.90}&\multirow{2}{*}{36.93} & \multirow{2}{*}{-27.46} \\ 
			& 30.33 & 36.08 & 11.62 & 35.75 & 37.50 & 3.48 & 44.98 & 63.46  \\ \cdashline{1-12}

            \multirow{2}{*}{\makecell[c]{$\text{MoELoRA}^{*}$}} & 80.15 & 49.60 & 96.65 & 58.40 & 51.54 & 22.22 & 65.79 & 60.10 & \multirow{2}{*}{33.73}& \multirow{2}{*}{39.32}  & \multirow{2}{*}{-26.83}  \\ 
& 67.15 & 39.17 & 4.87 & 33.78 & 25.31 & 0.71 & 38.74 & 60.10  \\ \cdashline{1-12} 
            
        \multirow{2}{*}{\makecell[c]{MoELoRA}} & 75.78 & 51.73 & 96.70 & 59.42 & 58.88 & 37.50 & 64.22& 60.08 & \multirow{2}{*}{\textcolor{blue}{\textbf{37.13}}}& \multirow{2}{*}{\textcolor{blue}{\textbf{42.76}}}  & \multirow{2}{*}{\textcolor{blue}{\textbf{-25.91}}}  \\ 
& 63.09 & 38.63 & 10.50 & 37.38 & 43.62 & 0.59 & 43.15 & 60.08  \\ \cdashline{1-12}

        \multirow{2}{*}{\makecell[c]{\textbf{BranchLoRA(Ours)}}} & 86.70 & 56.14 & 96.46 & 56.04 & 59.43 & 39.48 & 65.02 & 62.14 & \multirow{2}{*}{\textcolor{deepred}{\textbf{44.20}}}& \multirow{2}{*}{\textcolor{deepred}{\textbf{49.94}}}  & \multirow{2}{*}{\textcolor{deepred}{\textbf{-20.98}}}  \\ 
& 68.24 & 40.18 & 24.60 & 41.40 & 49.83 & 15.94 & 51.23 & 62.14  \\

\cdashline{1-12} 

   Multi-task & 56.77 & 49.35 & 95.55 & 56.65 & 53.90 & 30.09 & 59.50 & 55.65 & -- & 57.18 & -- \\

\bottomrule[1.2pt]

		\end{tabular}
	}
\caption{Main results on the LLaVA-1.5-7B model using the CoIN benchmark. For sequential finetuning methods (except for Zero-shot and Multi-task), the first row presents the results for each task evaluated immediately after tuning on the corresponding task (\emph{i.e.}, $A_{i,i}$), and the second row shows the results for each task after finetuning on the final task (\emph{i.e.}, $A_{T,i}$). The \textcolor{deepred}{\textbf{red}} highlights the highest overall performance, and the \textcolor{blue}{\textbf{blue}} indicates the second-highest performance. $*$ represents results from our re-implementation. Other results are cited from CoIN \cite{chen2024coin}.}
\label{main_results_llava}
\end{table*}

\subsection{Baselines} 

We evaluate the performance of BranchLoRA by comparing it against the following baselines: \textbf{Zero-shot}: Directly assessing each task using pre-trained MLLMs without additional finetuning; \textbf{LoRA} \cite{hulora}: Updating knowledge sequentially through two low-rank matrices, while preserving the original parameters of the pre-trained MLLM; \textbf{MoELoRA} \cite{chen2024coin}: Utilizing multiple identical yet independent LoRAs to capture specialized knowledge from sequential tasks and achieving State-Of-The-Art (SOTA) MCIT performance on the CoIN benchmark; \textbf{Multi-task}: Performing finetuning with LoRA on all tasks simultaneously, rather than using sequential training.

Moreover, we compare BranchLoRA with classic continual learning methods: \textbf{LwF} \cite{li2017learning} and \textbf{EWC} \cite{kirkpatrick2017overcoming}. More details about the above baselines can be found in Appendix \ref{ap:baselines}.

\subsection{Evaluation Metrics} 

We evaluate the outputs of MLLMs by comparing them to ground truths in a word-by-word manner. Since tasks produce outputs in various formats, the evaluation metrics are tailored accordingly. Detailed descriptions of these comparisons are provided in Appendix \ref{ap:comprison}.

In line with the CoIN benchmark \cite{chen2024coin}, we evaluate MCIT performance using three metrics: Average Accuracy (ACC) to measure performance after training on the final task, Mean Average Accuracy (MAA) to assess performance throughout the training process, and Backward Transfer (BWT) to quantify the extent of CF. These metrics are defined as follows: 
(1) $\text{ACC}=\frac{1}{T} \sum_{i=1}^T A_{T,i}$, where $A_{T,i}$ is the performance on $i$-th task after training the final task $T$. (2) $\text{MAA}=\frac{1}{T} \sum_{i=1}^T (\frac{1}{i} \sum_{k=1}^i A_{i,k})$, 
where $A_{i,k}$ is the performance on $k$-th task after training the  task $i$. (3) $\text{BWT} =\frac{1}{T} \sum_{i=1}^{T}(A_{T,i}-A_{i,i}),$
where $A_{i, i}$ is the performance on $i$-th task after training on $i$-th task.

\subsection{Implementation Details}

We use the well-established \textbf{LLaVA-1.5-7B} and \textbf{LLaVA-1.5-13B} \cite{liu2023llava} as our backbone models, integrating LoRA \cite{hulora} into the MLLM. 
During the MCIT process, both the vision encoder and the LLM remain frozen, with only the projector and LoRA components being finetuned. 
For a single LoRA, the rank $r$ is set to $128$, and the hyperparameter $\alpha$ is set to $256$. 
In the case of MoELoRA, we set the number of experts $N$ to $8$, and the rank of the small matrices within each expert is adjusted to $r/N = 128/8 = 16$ to ensure computational efficiency, as specified in its original paper. 
For a fair comparison, the number of experts and the rank in BranchLoRA are kept consistent with MoELoRA, selecting the top-$2$ experts, with the loss coefficient $\lambda$ set to $1.0$. 
All experiments are carried out on $8$ NVIDIA H800 GPUs, each with $80$GB of memory.

\begin{figure}[tbp]
\centering
  \includegraphics[width=1.0\linewidth]{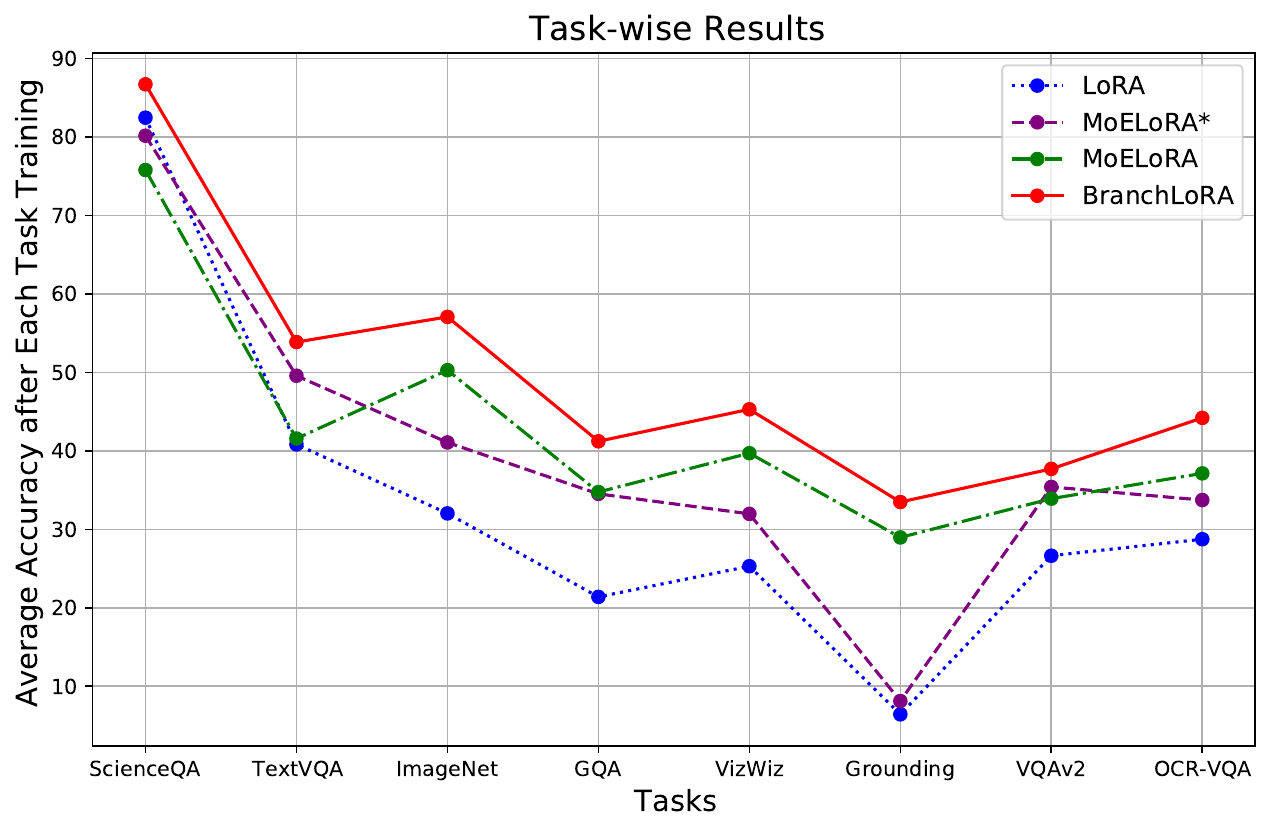}
    \caption{Task-wise performance comparison on the LLaVA-1.5-7B model. Our BranchLoRA method consistently outperforms the previous MCIT baselines, LoRA and MoELoRA, in all task-wise evaluations.}
    \label{fig:step}
\end{figure}

\section{Experimental Results}

\subsection{Main Results}\label{o_main_res}

To evaluate the performance of BranchLoRA, we conduct experiments on LLaVA-1.5-7B using the CoIN benchmark, with results shown in Table \ref{main_results_llava}.

Despite LLaVA-1.5-7B's vast knowledge, its Zero-shot performance on specialized tasks is suboptimal, with an MAA of $7.12$. While updating the model across all tasks simultaneously is resource-intensive, the Multi-task approach proves effective, reaching an MAA of $57.18$. 
Sequential finetuning methods like LoRA, MoELoRA, and BranchLoRA, outperform Multi-task on most tasks (\emph{i.e.}, the first row). This advantage may result from the model's ability to focus on one task, thereby minimizing interference caused by diverse instructions from all tasks. However, LoRA lacks strategies to mitigate CF, which leads to a loss of previously learned instructions and results in a BWT of $-32.62$.

MoELoRA alleviates CF by using distinct experts for each task, achieving ACC, MAA, and BWT scores of $37.13$, $42.76$, and $-25.91$, respectively. 
In contrast, BranchLoRA introduces an asymmetric structure that better balances shared and task-specific learning. 
It uses a flexible tuning-freezing mechanism and task-specific routers, allowing experts to capture intra-task knowledge, encourage cross-task collaboration, and reduce inter-task interference. 
This design significantly improves CF mitigation. 
Consequently, BranchLoRA outperforms MoELoRA by a substantial margin, achieving new SOTA performance on CoIN, with ACC, MAA, and BWT scores of $44.20$, $49.94$, and $-20.98$, respectively. 
Moreover, BranchLoRA significantly outperforms traditional continual learning methods like LwF and EWC.

As shown in Figure \ref{fig:step}, task-wise comparison results (\emph{i.e.}, task-wise MAA, which measures the average accuracy on all previous tasks, including the current one, after finetuning each task) further highlight the effectiveness of BranchLoRA. 
These results reveal that BranchLoRA consistently outperforms prior SOTA baselines, LoRA and MoELoRA, across all comparisons, showcasing its robustness and adaptability in sequential learning.

% \subsection{Task-wise Results}

% The notable improvement over LoRA and MoELoRA underscores BranchLoRA’s potential to set a new standard for the CoIN benchmark.

  \begin{table}[t]
  \centering
  \resizebox{1.0\linewidth}{!}{
  \begin{tabular}{lccccc}
  \toprule[1.2pt]
 \textbf{Method} & \makecell{\#. Trainable\\\ \ Parameters}$\Downarrow$ & \makecell{Training Time\\(ms/batch)}$\Downarrow$	& ACC$\Uparrow$ & MAA$\Uparrow$ & BWT$\Uparrow$ \\
  \hline
  MoELoRA & 350M&	62	& 37.13	& 42.76	& -25.91\\
  
  \textbf{BranchLoRA(Ours)} & \textbf{222M}	& \textbf{51}	& \textbf{44.20}	& \textbf{49.94}	& \textbf{-20.98}
\\
  \bottomrule[1.2pt]
  \end{tabular}}
  \caption{The efficiency analysis of BranchLoRA on LLaVA-1.5-7B. The \textbf{Bold} represents the best results.}
  \label{efficiency}
  \end{table}

  \begin{table}[t]
  \centering
  \resizebox{1.0\linewidth}{!}{
  \begin{tabular}{lccc}
 \toprule[1.2pt]
 \textbf{Variant} & ACC$\Uparrow$ & MAA$\Uparrow$ & BWT$\Uparrow$ \\
\hline
  MoELoRA            & 37.13  & 42.76    & -25.91  \\
  \ \ \ +\ shared matrix $\bm{A}$            & 38.19 & 43.95    & -25.32  \\
  \ \ \ +\ dynamic sparse selection strategy  & 39.96 & 45.53    & -23.77  \\
 \ \ \ +\ flexible tuning-freezing mechanism  & 42.22 & 47.76    & -22.41  \\
  \ \ \ +\ task-specific router (\textbf{BranchLoRA}) & \textbf{44.20} & \textbf{49.94}    & \textbf{-20.98}  \\
  \bottomrule[1.2pt]
  \end{tabular}}
  \caption{The ablation study of BranchLoRA on LLaVA-1.5-7B. When compared with BranchLoRA, all ablation variants degrade MCIT performance. It verifies the importance of all components to address MCIT collaboratively. The \textbf{Bold} represents the best results.}
  \label{ablation}
  \end{table}

\begin{table*}[t]
	\renewcommand\arraystretch{1.3}
	\renewcommand\tabcolsep{4.0pt}
	\centering
	\resizebox{\linewidth}{!}{
		\begin{tabular}{ccccccccc|ccc}
			\toprule[1.2pt]
			\multirow{2}{*}{\textbf{Method}} &
			\multicolumn{8}{c|}{\textbf{Accuracy on Each Task}} &
			\multicolumn{3}{c}{\textbf{Overall Results}} \\ \cline{2-12}
			& ScienceQA & TextVQA & ImageNet & GQA & VizWiz & Grounding & VQAv2 & OCR-VQA &ACC$\Uparrow$ & MAA$\Uparrow$ & BWT$\Uparrow$  \\ 
			\hline

			\multirow{2}{*}{MoELoRA} &  76.16	& 55.92	& 97.67	& 55.85&	62.95	& 48.10	& 68.32	& 64.05	 & \multirow{2}{*}{{42.51	}} & \multirow{2}{*}{{ 49.14}}  & \multirow{2}{*}{	-23.62}  \\ 
&70.66	& 45.82	& 12.90	& 36.15	& 49.93	& 10.79	& 49.76	& 64.05 \\ 
 
\cdashline{1-12}

			\multirow{2}{*}{BranchLoRA} & 87.26 & 60.97 & 	98.28 & 54.08 & 63.89 & 46.98 & 70.61 & 66.37 & \multirow{2}{*}{\textbf{49.27}}& \multirow{2}{*}{\textbf{55.73}}  & \multirow{2}{*}{\textbf{-19.29}}  \\ 
            
& 78.06	& 49.85	& 29.46	& 39.76	& 55.28	& 19.01	& 56.36 &	66.37  \\ 

\bottomrule[1.2pt]
		\end{tabular}
	}
    \caption{The results of MoELoRA and BranchLoRA on the larger LLaVA-1.5-13B model.}
\label{backbone_size}
\end{table*}

\begin{table*}[t]
	\renewcommand\arraystretch{1.3}
	\renewcommand\tabcolsep{4.0pt}
	\centering
	\resizebox{\linewidth}{!}{
		\begin{tabular}{ccccccccc|ccc}
			\toprule[1.2pt]
			\multirow{2}{*}{\textbf{Type}} &
			\multicolumn{8}{c|}{\textbf{Accuracy on Each Task}} &
			\multicolumn{3}{c}{\textbf{Overall Results}} \\ \cline{2-12}
			& ScienceQA & TextVQA & ImageNet & GQA & VizWiz & Grounding & VQAv2 & OCR-VQA & ACC$\Uparrow$ & MAA$\Uparrow$ & BWT$\Uparrow$  \\ 
			\hline
			\multirow{2}{*}{Original} & 86.70 & 56.14 & 96.46 & 56.04 & 59.43 & 39.48 & 65.02 & 62.14 & \multirow{2}{*}{44.20}& \multirow{2}{*}{49.94}  & \multirow{2}{*}{-20.98}  \\ 
& 68.24 & 40.18 & 24.60 & 41.40 & 49.83 & 15.94 & 51.23 & 62.14  \\ 
 
 \cdashline{1-12}
			
			\multirow{2}{*}{Diverse} & 86.70 & 57.26 & 97.42 & 54.87 & 56.46 & 37.94 & 67.91 & 64.16 & \multirow{2}{*}{45.06}& \multirow{2}{*}{50.25}  & \multirow{2}{*}{-20.28}  \\ 
& 69.52 & 43.39 & 23.84 & 44.71 & 46.84 & 15.30 & 52.74 & 64.16  \\ 

\cdashline{1-12}
			
			\multirow{2}{*}{10Type} & 88.43 & 58.98 & 98.30 & 55.97 & 54.77 & 39.21 & 69.52 & 65.44 & \multirow{2}{*}{\textbf{46.47}} & \multirow{2}{*}{\textbf{51.76}}  & \multirow{2}{*}{\textbf{-19.86}}  \\ 
& 70.22 & 44.82 & 25.37 & 43.82 & 47.67 & 19.57 & 54.82 & 65.44  \\ 
            
            \bottomrule[1.2pt]
		\end{tabular}
	}
	\caption{The results of BranchLoRA on the LLaVA-1.5-7B model about different instruction templates.}
	\label{instruction_type}
\end{table*}

\subsection{Efficiency Analysis}

We compare the number of trainable parameters and training time of BranchLoRA with the previous SOTA, MoELoRA, with results presented in Table \ref{efficiency} (based on LLaVA-1.5-7B). 
Training time refers to the duration required for forward and backward propagation of a data batch, measured in milliseconds (ms). 
To minimize variance, we averaged the time over $100$ batches. 
BranchLoRA significantly reduces both trainable parameters and training time compared to MoELoRA, while maintaining superior performance, providing quantitative evidence of the efficiency of our approach.

\subsection{Ablation Study}

This section presents ablation studies on the LLaVA-1.5-7B model to evaluate the contributions of individual components in BranchLoRA, as detailed in Table~\ref{ablation}. 
Sharing multiple $\bm{A}$ matrices within MoELoRA not only significantly improves parameter efficiency but also slightly enhances MCIT performance, indicating that the asymmetric structure better balances task-shared and task-specific learning. Introducing a dynamic sparse selection strategy further improves performance by reducing the impact on previously learned knowledge compared to aggregating all experts. Building on this, the flexible tuning-freezing mechanism enhances intra-task learning while fostering cross-task collaboration, leading to additional performance improvements. Finally, the incorporation of task-specific routers minimizes inter-task interference, resolving the challenge of continuous updates causing expert distributions to overly favor the most recent task. Together, these components constitute the complete BranchLoRA framework. 

Moreover, despite occasional misclassifications by the automatic task selector during inference (achieving an average accuracy of $95.8$\% across all tasks), BranchLoRA consistently outperforms SOTA methods across various metrics. Its ability to automatically select tasks during inference aligns with the demands of real-world applications, further highlighting its practical value.

\subsection{More Explorations}

\paragraph{Larger Backbone Size}

We conduct experiments on larger MLLM backbones, such as LLaVA-1.5-13B. The results, presented in Table \ref{backbone_size}, highlight the scalability and effectiveness of BranchLoRA in resource-intensive scenarios. Larger models like LLaVA-1.5-13B experience less CF compared to smaller models like LLaVA-1.5-7B (shown in Table \ref{main_results_llava}) in MCIT across multiple tasks. However, CF still occurs, particularly when there is a significant difference in task similarity. Across both the 7B and 13B models, BranchLoRA consistently outperforms the previous SOTA, MoELoRA, demonstrating the scalability and versatility of our approach across different MLLM backbone sizes.

\paragraph{Impact of Instruction Diversity}

In our main experiment (Section \ref{o_main_res}), some tasks rely on similar instruction templates (\emph{i.e.}, Original). To examine how template variety affects the MCIT performance of our BranchLoRA, we introduce two additional template types: Diverse and 10Type. The Diverse template involves distinct instructions tailored to each task, while the 10Type template involves randomly selecting from $10$ different instruction templates for each task. A detailed list of these three instruction types for all tasks is provided in Table \ref{tab:diversity} of Appendix \ref{ap:template}.

Table \ref{instruction_type} shows the MCIT performance of our BranchLoRA across these three template types. Our analysis reveals that simply switching to the Diverse template type has a limited effect on overall performance. However, employing random selection from multiple instruction templates significantly improves overall results. This improvement likely stems from the model's ability to better understand the underlying instructional intent when exposed to varied templates. 

These findings highlight the value of incorporating diverse instructions for each task rather than relying on a single instruction. Introducing instruction variety enhances the model's capacity to interpret instructional intent, mitigates the decline in instruction-following ability, and improves robustness to variations in instructions.

\section{Conclusion}

In this paper, we present BranchLoRA, an innovative solution to address the critical parameter inefficiency and CF in the MoELoRA framework for MCIT. By introducing a flexible tuning-freezing mechanism and task-specific routers with automatic selector, BranchLoRA enables experts to specialize in intra-task knowledge while promoting inter-task collaboration, more effectively mitigating CF. Extensive experiments on different MLLM sizes (\emph{e.g.}, LLaVA-1.5-7B and LLaVA-1.5-13B) using the latest CoIN benchmark demonstrate that BranchLoRA significantly outperforms MoELoRA, offering a more efficient and robust approach for continual alignment with human intent across sequential tasks.

In the future, a promising direction is to integrate BranchLoRA with advanced model merging techniques \cite{alexandrov2024mitigating,merge1,merge2,merge3,merge4,merge5}, which typically assign different levels of importance to task-specific features. These methods enable more nuanced consolidation of knowledge across tasks. Incorporating them into the BranchLoRA framework could further mitigate CF by dynamically preserving essential information from previous tasks while adapting to new ones. This synergy has the potential to improve the robustness and scalability of BranchLoRA in more diverse and extended MCIT settings.

\section*{Limitations}

Despite its promising results, BranchLoRA has some limitations that need to be addressed. For instance, our experiments were primarily conducted using the recent MCIT benchmark, which may not fully capture the method's potential across diverse tasks and domains. To further validate the effectiveness of BranchLoRA, future research could explore its application on a wider array of continual instruction tuning benchmarks, including those that focus on non-multimodal tasks. Additionally, developing more comprehensive and challenging benchmarks that encompass a broader variety of multimodal tasks, real-world scenarios, and different domains would provide a more robust evaluation of BranchLoRA's performance. Such efforts would help assess the method’s scalability, adaptability, and generalization capabilities in a broader context, enabling a deeper understanding of its strengths and limitations.

\bibliography{custom}

% \newpage

\appendix

\begin{table*}[t]
	\centering
	\resizebox{\linewidth}{!}{
		\begin{tabular}{cccccc}
			\toprule[1.2pt]
			\textbf{Task} &
			\textbf{Dataset} &
			\textbf{Instruction} &
			\makecell[c]{\textbf{Train} \\ \textbf{Number}}
			& \makecell[c]{\textbf{Test} \\ \textbf{Number}} \\
			\midrule

        \makecell[c]{\textbf{Knowledge Grounded IQA}} & ScienceQA & \makecell[c]{Answer with the option's letter \\ from the given choices directly} &  12k & 4k \\ \hdashline

        \makecell[c]{\textbf{Reading Comprehension IQA}} & \makecell[c]{TextVQA} & \makecell[c]{Answer the question using a \\ single word or phrase} &  34k & 5k \\ \hdashline

		\multicolumn{1}{c}{\textbf{Classification}} & ImageNet & \makecell[c]{What is the object in the image? \\ Answer the question using a \\ single word or phrase} &  129k & 5k \\ \hdashline

		\makecell[c]{\textbf{Visual Reasoning IQA}} & GQA & \makecell[c]{Answer the question using a \\ single word or phrase} &  72k & 1k \\ \hdashline

\makecell[c]{\textbf{Blind People IQA}} & VizWiz & \makecell[c]{Answer the question using a \\ single word or phrase} &  20k & 8k \\ \hdashline

			\multicolumn{1}{c}{\makecell[c]{\textbf{Grounding}}} & \makecell[c]{RefCOCO\;\;\; \\ RefCOCO+ \\ RefCOCOg\;} & \makecell[c]{Please provide the bounding \\ box coordinate of the region \\ this sentence describes: <description>} &  55k & 31k \\ \hdashline

			\textbf{Image Question Answering (IQA)} & VQAv2 & \makecell[c]{Answer the question using a \\ single word or phrase} &  82k & 107k \\ \hdashline

			\makecell[c]{\textbf{OCR IQA}} & OCR-VQA & \makecell[c]{Answer the question using a \\ single word or phrase} &  165k & 100k \\ 
			\bottomrule[1.2pt]
		\end{tabular}
	}
\caption{The statistic of collected multimodal datasets in the CoIN benchmark \cite{chen2024coin}.}
	\label{aptable:datasets}
\end{table*}

\section{Datasets}
\label{ap:datasets}
The detailed statistics for the eight multimodal datasets included in the CoIN benchmark \cite{chen2024coin} are presented in Table \ref{aptable:datasets}.

\section{Baselines}
\label{ap:baselines}

We assess the performance of BranchLoRA by comparing it with several baselines: \textbf{Zero-shot}: Evaluating each task directly with pre-trained MLLMs, without any further finetuning; \textbf{LoRA} \cite{hulora}: Updating knowledge sequentially through two low-rank matrices, while retaining the original parameters of the pre-trained MLLM; \textbf{MoELoRA} \cite{chen2024coin}: Using multiple identical yet independent LoRAs to capture specialized knowledge from sequential tasks, achieving SOTA performance on the CoIN benchmark; \textbf{Multi-task}: Performing finetuning with LoRA on all tasks simultaneously, rather than using sequential training.

Additionally, we perform comparisons with classic continual learning techniques, such as \textbf{LwF} \cite{li2017learning} and \textbf{EWC} \cite{kirkpatrick2017overcoming}. 
Following the CoIN benchmark \cite{chen2024coin}, we compute the Fisher matrix for EWC by randomly selecting $1000$ samples from each task and set the hyperparameter $\lambda$ to $0.1$. 
For LwF, we retain $100$ logits per task to calculate the distillation loss, with $\lambda$ also set to $0.1$.

\section{Comparison Details}\label{ap:comprison}

In line with the CoIN benchmark \cite{chen2024coin}, we evaluate performance on the Image Question Answering task (encompassing VQAv2, ScienceQA, TextVQA, GQA, VizWiz, and OCR-VQA) by measuring the accuracy of predicted answers against the ground truth, similar to the approach used in LLaVA \cite{liu2023llava}. For classification tasks, the evaluation metric involves comparing the predicted labels to the actual ones. In the referring expression comprehension (grounding) task, we adopt the commonly used Intersection over Union (IoU) metric to assess prediction accuracy. A prediction is deemed correct if the IoU between the predicted and ground-truth bounding boxes exceeds $0.5$.

\begin{table*}[t]
\centering
\resizebox{\linewidth}{!}{
\begin{tabular}{cccc}
\toprule[1.2pt]
\textbf{Task} &
\textbf{Original} &
\textbf{Diverse} &
\textbf{10Type} \\
\hline

\multicolumn{1}{c}{\textbf{ScienceQA}} & \makecell[c]{Answer with the option’s \\ letter from the given \\ choices directly} & \makecell[c]{Answer with the option’s \\ letter from the given \\ choices directly} & \makecell[l]{Answer with the option’s letter from the given choices directly \\ Select the correct answer from the given choices and respond with the letter of the chosen option \\ Determine the correct option from the provided choices and reply with its corresponding letter \\ Pick the correct answer from the listed options and provide the letter of the selected option \\ Identify the correct choice from the options below and respond with the letter of the correct option \\ From the given choices, choose the correct answer and respond with the letter of that choice \\ Choose the right answer from the options and respond with its letter \\ Select the correct answer from the provided options and reply with the letter associated with it \\ From the given choices, select the correct answer and reply with the letter of the chosen option \\ Identify the correct option from the choices provided and respond with the letter of the correct option \\ From the given choices, pick the correct answer and respond by indicating the letter of the correct option}  \\ \hdashline

\multicolumn{1}{c}{\textbf{TextVQA}} & \makecell[c]{Answer the question \\  using a single \\ word or phrase} & \textcolor{CoINRed}{\makecell[c]{Capture the essence of your \\ response in a single word \\ or a concise phrase}} & \makecell[l]{Answer the question with just one word or a brief phrase \\ Use one word or a concise phrase to respond to the question \\ Answer using only one word or a short, descriptive phrase \\ Provide your answer in the form of a single word or a brief phrase \\ Use a single word or a short phrase to respond to the question \\ Summarize your response in one word or a concise phrase \\ Respond to the question using a single word or a brief phrase \\ Provide your answer in one word or a short, descriptive phrase \\ Answer the question with a single word or a brief, descriptive phrase \\ Capture the essence of your response in one word or a short phrase \\ Capture the essence of your response in a single word or a concise phrase} \\ \hdashline

\multicolumn{1}{c}{\textbf{ImageNet}} & \makecell[c]{Answer the question \\  using a single \\ word or phrase} & \textcolor{CoINRed}{\makecell[c]{Express your answer in \\ a single word or a \\ short, descriptive phrase}} & \makecell[l]{Express your answer in a single word or a short, descriptive phrase \\ Provide your answer using a single word or a brief phrase \\ Describe the content of the image using one word or a concise phrase \\ Respond to the question with a single word or a short, descriptive phrase \\ Classify the image content using only one word or a brief phrase \\ Give your answer in the form of a single word or a concise phrase \\ Use a single word or a short phrase to categorize the image content \\ Express your answer with one word or a short, descriptive phrase \\ Identify the type of content in the image using one word or a concise phrase \\ Summarize your response in a single word or a brief phrase \\ Use one word or a short phrase to classify the content of the image} \\ \hdashline

\multicolumn{1}{c}{\textbf{GQA}} & \makecell[c]{Answer the question \\  using a single \\ word or phrase} & \textcolor{CoINRed}{\makecell[c]{Respond to the question \\ briefly, using only one \\ word or a phrase}} & \makecell[l]{Respond to the question with a single word or a short phrase \\ Respond to the question using only one word or a concise phrase \\ Answer the question with a single word or a brief phrase \\ Respond with one word or a short phrase \\ Provide your answer in the form of a single word or a concise phrase \\ Respond to the question with just one word or a brief phrase \\ Answer the question using a single word or a concise phrase \\ Provide your response using only one word or a short phrase \\ Respond to the question with a single word or a brief phrase \\ Respond to the question using just one word or a concise phrase \\ Answer the question with one word or a short phrase} \\ \hdashline

\multicolumn{1}{c}{\textbf{VizWiz}} & \makecell[c]{Answer the question \\  using a single \\ word or phrase} & \textcolor{CoINRed}{\makecell[c]{Provide a succinct \\ response with a single \\ word or phrase}} & \makecell[l]{Answer the question using only one word or a concise phrase \\ Respond to the question using only one word or a concise phrase \\ Respond to the question with a single word or a brief phrase \\ Provide your answer using just one word or a short phrase \\ Respond with one word or a concise phrase \\ Answer the question with just one word or a brief phrase \\ Use a single word or a short phrase to answer the question \\ Provide your answer in the form of one word or a brief phrase \\ Reply to the question using one word or a concise phrase \\ Answer with a single word or a short phrase \\ Use one word or a brief phrase to answer the question} \\ \hdashline

\multicolumn{1}{c}{\textbf{Grounding}} & \makecell[c]{Please provide the bounding \\ box coordinate of the region \\ this sentence describes} & \makecell[c]{Please provide the bounding \\ box coordinate of the region \\ this sentence describes} & \makecell[l]{Identify and provide the bounding box coordinates that match the description given in this sentence \\ Extract and provide the bounding box coordinates based on the region described in the sentence \\ Please provide the bounding box coordinate of the region this sentence describes \\ Find and provide the bounding box coordinates for the region mentioned in the sentence \\ Provide the coordinates of the bounding box that correspond to the region described in the sentence \\ Give the bounding box coordinates as described in the sentence \\ Determine and provide the bounding box coordinates based on the description in the sentence \\ Identify and provide the coordinates of the bounding box described in the sentence \\ Provide the coordinates for the bounding box based on the region described in the sentence \\ Extract and provide the coordinates for the bounding box described in the sentence \\ Identify and give the coordinates of the bounding box as described by the sentence} \\ \hdashline

\multicolumn{1}{c}{\textbf{VQAv2}} & \makecell[c]{Answer the question \\  using a single \\ word or phrase} & \makecell[c]{Answer the question \\  using a single \\ word or phrase} & \makecell[l]{Answer the question using a single word or phrase \\ Answer the question with a single word or a brief phrase \\ Use one word or a short phrase to respond to the question \\ Answer the question using just one word or a concise phrase \\ Provide your answer to the question using only one word or a brief phrase \\ Respond to the question with a single word or a short phrase Use a single word or phrase to answer the question \\ Provide an answer using only one word or a brief phrase \\ Answer the question succinctly with one word or a brief phrase \\ Answer the question with just one word or a short phrase \\ Respond to the question using a single word or a concise phrase} \\\hdashline

\multicolumn{1}{c}{\textbf{OCR-VQA}} & \makecell[c]{Answer the question \\  using a single \\ word or phrase} & \textcolor{CoINRed}{\makecell[c]{Condense your answer for \\ each question into a single \\  word or concise phrase}} & \makecell[l]{Answer with the option's letter from the given choices directly \\ Select the correct answer from the given choices and respond with the letter of the chosen option \\ Determine the correct option from the provided choices and reply with its corresponding letter \\ Pick the correct answer from the listed options and provide the letter of the selected option \\ Identify the correct choice from the options below and respond with the letter of the correct option \\ From the given choices, choose the correct answer and respond with the letter of that choice \\ Choose the right answer from the options and respond with its letter \\ Select the correct answer from the provided options and reply with the letter associated with it \\ From the given choices, select the correct answer and reply with the letter of the chosen option \\ Identify the correct option from the choices provided and respond with the letter of the correct option \\ From the given choices, pick the correct answer and respond by indicating the letter of the correct option} \\

\bottomrule[1.2pt]
\end{tabular}}
\caption{The list of different instruction templates for each task \cite{chen2024coin}.}
\label{tab:diversity}
\end{table*}

\section{Different Instruction Templates}\label{ap:template}

The list of instruction templates for each task is provided in Table \ref{tab:diversity}. \textbf{Original}: Certain tasks use similar instructions. \textbf{Diverse}: Unique instruction templates specifically designed for each task. \textbf{10Type}: A randomly selected instruction template from a set of $10$ distinct templates for each task.

\end{document}